\documentclass{llncs}
\usepackage{cite}  % allows for indexgeneration
\usepackage{graphicx}
\usepackage{amsmath}
\usepackage{amsfonts}
\usepackage{subfig}

\begin{document}
	
\title{Adversarial Image Registration with Application for MR and TRUS Image Fusion}
\author{Pingkun Yan\inst{1}, Sheng Xu\inst{2}, Ardeshir R. Rastinehad\inst{3}, Brad J. Wood\inst{2}}
\institute{Department of Biomedical Engineering, Rensselaer Polytechnic Institute, Troy NY 12180 \and
	National Institutes of Health, Center for Interventional Oncology, Radiology \& Imaging Sciences, Bethesda, MD 20892 \and
	Icahn School of Medicine at Mount Sinai, New York, NY 10029
}

\maketitle 

\begin{abstract}
Robust and accurate alignment of multimodal medical images is a very challenging task, which however is very useful for many clinical applications. For example, magnetic resonance (MR) and transrectal ultrasound (TRUS) image registration is a critical component in MR-TRUS fusion guided prostate interventions. However, due to the huge difference between the image appearances and the large variation in image correspondence, MR-TRUS image registration is a very challenging problem. In this paper, an adversarial image registration (AIR) framework is proposed. By training two deep neural networks simultaneously, one being a generator and the other being a discriminator, we can obtain not only a network for image registration, but also a metric network which can help evaluate the quality of image registration. The developed AIR-net is then evaluated using clinical datasets acquired through image-fusion guided prostate biopsy procedures and promising results are demonstrated.
\end{abstract}

% ======= ======= ======= ======= %

\section{Introduction}

Prostate cancer is one of the leading causes of cancer death among men in the western world. The fusion of magnetic resonance (MR) and transrectal ultrasound (TRUS) images, benefited by the good sensitivity and specificity of multi-parametric MR (mpMR) on identifying suspicious prostate cancer regions, has been demonstrated improving the biopsy yield by as much as 30\% \cite{siddiqui_comparison_2015}. For a fusion system to work effectively, accurate registration of different imaging modalities is critical. However, multi-modality image registration is a very challenging task, as it is hard to define a robust image similarity metric \cite{Cao2016}. The registration of MR and TRUS is more difficult due to the noisy appearance of ultrasound images and the inhomogeneous imaging resolutions between MR and TRUS.

With the rapid advancement of deep learning technology in the past several years, a number of new image registration methods based on deep learning have been proposed, which gained better performance compared to the traditional methods. The early deep learning based image registration methods still follow the classical framework of iteratively optimizing over certain similarity metric through updating the transformation. Deep learning was initially only used for acquiring a better similarity metric. For example, Cheng et al. \cite{Cheng_deepSimilarity_2015} used a multilayer perceptron network to learn the correspondence between a pair of images. Simonovsky et al. \cite{Simonovsky_2016_miccai} developed a convolutional neural network (CNN) based similarity learning network and embedded it into an image registration framework for multi-modal image alignment. Compared with the traditional manually defined similarity measures like mutual information, deep learning similarity metric uses huge number of automatically extracted features to achieve better performance. Its output value can also provide a good sense of the registration quality due to the pre-defined value range.

With more powerful CNN being designed to extract more representative image features, Miao et al. \cite{miao_cnn_2016} proposed a CNN based method to directly estimate the transformation parameters instead of using an iterative process. Therefore, the registration can be performed very fast and efficient. De Vos et al. \cite{de_vos_end--end_2017} further developed an end-to-end unsupervised registration method, which however is limited to same modality image registration. Recently, Hu et al. \cite{hu_label-driven_2017} proposed a label-driven registration method by using CNN to evaluate not only image pairs but also the object label pairs for MR-TRUS image registration. Cao et al. \cite{cao_deep_2018} developed a deep learning method for inter-modality image registration without using ground truth but supervised by intra-modality similarity. However, such methods lack a direct feedback of registration quality, which can be important for image-fusion guided interventions.

\begin{figure}[tb]
	\centering
	\includegraphics[width=\textwidth]{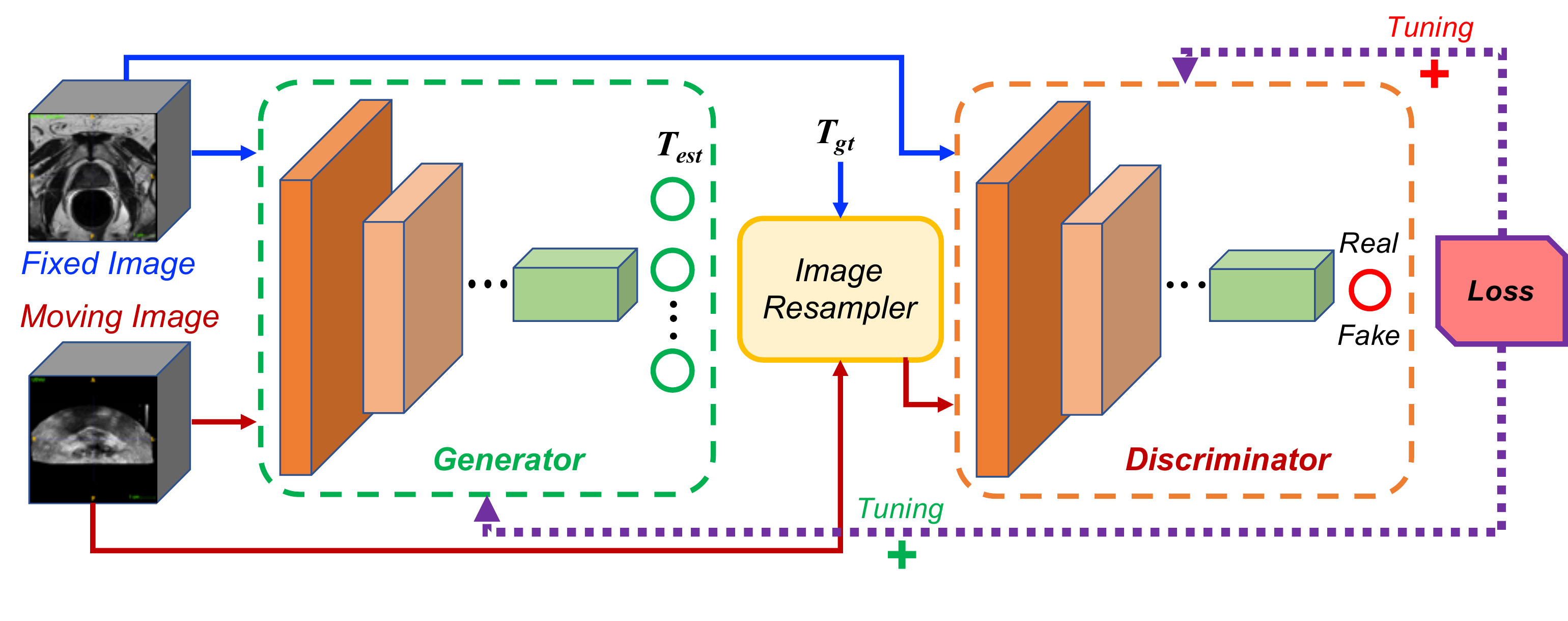}
	\caption{\label{fig:framework}Overall structure of the proposed AIR-net registration framework.}
\end{figure}

%Inexact registration ground truth for training

Inspired by the previous works, in this paper, we propose a multi-modality image registration method based on the generative adversarial network (GAN) framework \cite{goodfellow_generative_2014} with simultaneously trained CNNs for transformation parameter estimation and registration quality evaluation. The proposed adversarial image registration network (AIR-net) consists of two sub-networks, registration generator and registration discriminator, which are trained in the adversarial fashion. An overview of the proposed AIR-net is shown in Fig.~\ref{fig:framework}. 

In the proposed method, the registration generator network (\textbf{G}) directly estimates transformation parameters between the input image pair. The image resampler then uses either the estimated transformation $T_{est}$ or the ground truth transformation $T_{gt}$ to interpolate the input moving image to get a new resampled moving image. The registration discriminator (\textbf{D}) tries to tell if its input image pair is aligned using transformation $T_{est}$ or $T_{gt}$. As the training goes on, both \textbf{G} and \textbf{D} are iteratively updated. The feedback of \textbf{D} will be used to improve \textbf{G}, so that eventually \textbf{G} will be well trained to generate transformations close to $T_{gt}$ to pass the test of \textbf{D}. 

Our work in this paper has two major contributions. First, the proposed AIR-net not only estimates transformation parameters directly with an efficient feed-forward pass of G-network but also evaluates the quality of the estimated registration with the D-network, which makes it very suitable for applications like image-guided intervention. Second, the AIR-net is trained in an end-to-end fashion, where both \textbf{G} and \textbf{D} become available once the training is completed. Our experimental results demonstrate the effectiveness of the proposed approach.

The rest of this paper is organized as follows. Section~\ref{sec:AIR} gives details of the proposed AIR-net. The network training and experimental results are presented in Section~\ref{sec:exp}. Finally, Section~\ref{sec:conclusions} draws conclusions.

% =================================== %
\section{Adversarial Image Registration (AIR)}
\label{sec:AIR}

\subsection{Generator and Discriminator Networks}

In our work, the G- and D-networks are designed using CNNs due to their strong capability for image feature extraction and compact representation. The MR and TRUS volumes in our work are 3D data. However, to build deep CNNs to effectively deal with the complex nature of this challenging multi-modality image registration problem, we consider each 3D volume as multi-channel 2D image. In this way, much deeper neural networks can be trained on a single GPU with limited memory compared with using 3D CNNs. We also experimented with 3D CNNs with shallower structures, and our results showed that deeper 2D CNNs indeed performed better.

The structure of the designed G-network is as follows.  It first starts with a dilated convolutional layer, aka atrous convolution, to enlarge the perceptive field. The layer has 128 filters with dilation of 2. All the convolutional filters in the designed networks are in the size of 3$\times$3, if not explicitly noted. Each convolutional layer is followed by a rectified linear unit (ReLU) layer as activation. The first convolutional layer is followed by two more convolutional layers with 128 filters and stride of 2 to reduce the output tensor size. After that, a residual block containing 3 convolutional layers with residual connections as in \cite{he_deep_2016} is used to have both high- and low-level features. The number of filters remains to be 128. We then used a 1$\times$1 convolutional layer to decrease the number of filters from 128 to 8, in order to reduce the number of parameters. Two fully connected layers are then used to get the final output. The first fully connected layer has 256 hidden units, while the number of the units for the second one is equal to the transformation parameters, e.g. 6 for 3D rigid transformation and 12 for 3D affine transformation. There is no activation function for the output layer of G-network.
The D-network has almost identical structure as the G-network, except that the last fully connected layer has only one output unit with Sigmoid activation function, which is for evaluating the performance of registration.

The input to the networks is in the form of ``two-channel'' images, which are obtained by concatenating the MR and TRUS image pair. The choice is made based on the extensive experiments performed in \cite{ZagoruykoK15}, where CNN was used to compare image patches from natural images. We believe that the conclusion also applies to medical image registration, as confirmed by the work of Simonovsky et al. \cite{Simonovsky_2016_miccai}.

\subsection{Adversarial Training}

The designed networks can then be trained in the adversarial fashion. However, as it is known that the original GAN \cite{goodfellow_generative_2014} can be tricky to train due to the unstable loss, the improved version of Wasserstein GAN (WGAN) by Arjovsky et al. \cite{arjovsky_wasserstein_2017} is adopted in our work. To make the network quickly converge to generate good image registrations, perturbed transformations are also used to compute part of the loss so the networks can recognize poor registrations.
Let $I_f$ and $I_m$ denote the fixed image and the moving image, respectively, corresponding to MR image and TRUS image in this application. Assume that $I_m$ has been properly registered to $I_f$ by using the ground truth transformation. Then the discriminator loss $\mathcal{L}(D)$ is defined as
\begin{equation}
\mathcal{L}(D) = - \mathbb{E}_{T\sim p_{gt}(T)} [D(I_f, I_m)] + \mathbb{E}_{T\sim p_{z}(T)} [D(I_f, T(I_m)],
\end{equation}
where $\mathbb{E}_{T\sim p_{gt}(T)} [D(I_f, I_m)]$ denotes the error expectation of the discriminator given a well aligned MR-TRUS image pair and $\mathbb{E}_{T\sim p_{z}(T)} [D(I_f, T(I_m)]$ defines the error expectation of the discriminator given a randomly perturbed transformation. The generator loss $\mathcal{L}(G)$ is defined as
\begin{equation}
\mathcal{L}(G) = \mathbb{E}_{T\sim p_{z}(T)} [1 - D(I_f, T_{est} (T(I_m))) + \alpha \| T_{est} - T^{-1} \|^2], %\\
%
%\min_G\max_D L(D,G) &= \min_G\max_D \left[\mathcal{L}(D) + \mathcal{L}(G) + \| T_g - T_z \|^2 \right].
\end{equation}
where $T_{est}$ is the registration transform generated by the generator $G(I_f, T(I_m))$ and $\| T_{est} - T^{-1} \|^2$ is the Euclidean distance between the estimated transformation and the randomly created transformation. The latter is weighted by a positive weighting parameter $\alpha$. 

For WGAN, after each round of training, the parameters of the D-network needs to be clipped for stability. The clipping parameter was set to be 0.01 in our work. The G-network is trained once the D-network is updated twice, i.e. the parameter of critic is set to be 2.
It is worth noting that although we used the square of difference between the transformation parameters as part of the generator loss, the AIR-net can still be trained without it. The training process just takes longer and the parameters need to be tuned carefully.

% =================================== %
\section{Experiments}
\label{sec:exp}

%\subsection{Implementation}
The presented method is implemented in Python based on the PyTorch deep learning library \cite{pytorch}. To realize an end-to-end training of the network with resampling component in between of the two networks, the technique of spatial transform network proposed by Jaderberg et al. \cite{jaderberg_spatial_2015} is used. 

\subsection{Materials and Training}

%\begin{figure}[tb]
%	\centering
%	\includegraphics[width=.6\textwidth]{./figs/wgan2d_loss}
%	\caption{\label{fig:loss}Training loss curves of the proposed AIR-net for MR-TRUS image registration.}
%\end{figure}

In our work, a total 763 sets of data have been used for experiments, with 679 from the National Institutes of Health and the other 84 from the Mount Sinai Hospital. The data were acquired from MR-TRUS fusion-guided prostate cancer biopsy procedures using FDA approved UroNav device (In Vivo, FL, USA). Each case contains a T2-weighted MR volume, a 3D TRUS volume reconstructed from 2D ultrasound sweep of the prostate under electro-magnetic tracking. Each MR volume has 512$\times$512$\times$26 voxels with the resolution of 0.3mm$\times$0.3mm$\times$3mm. The ultrasound volumes have varying sizes and resolutions, which are determined by the ultrasound scanning parameters used during the procedure. The data were randomly split into training and validation sets with a ratio of 5:1, resulting in  636 cases for training and 127 cases for validation.

The MR and TRUS volumes are sampled into the size of 256$\times$256 multi-channel images. The perturbed transformation parameters are in the following ranges: rotation is in [-25,25] degrees and translation is in [-5,5]mm.

%Fig.~\ref{fig:loss} shows the curves of losses for the training process. It can be seen that as the training goes on, the loss of D-network $\mathcal{L}(D)$ starts to decrease, which means that the discriminator is getting better. It then causes the loss of generator $\mathcal{L}(G)$ to increase, as many transformations made by G are being recognized. The training converges after the two networks become stable.
%
The developed network is trained and tested on a workstation equipped with a NVIDIA Titan Xp GPU. It take about 8 hours for the network to get trained on our dataset. When testing on an image pair, it runs very fast, using less than 100ms for estimating a transformation. We then can use both the generator and discriminator networks efficiently to iteratively update the image registration until it converges.

\begin{figure}[tbh!]
	\centering
	\includegraphics[width=.9\textwidth]{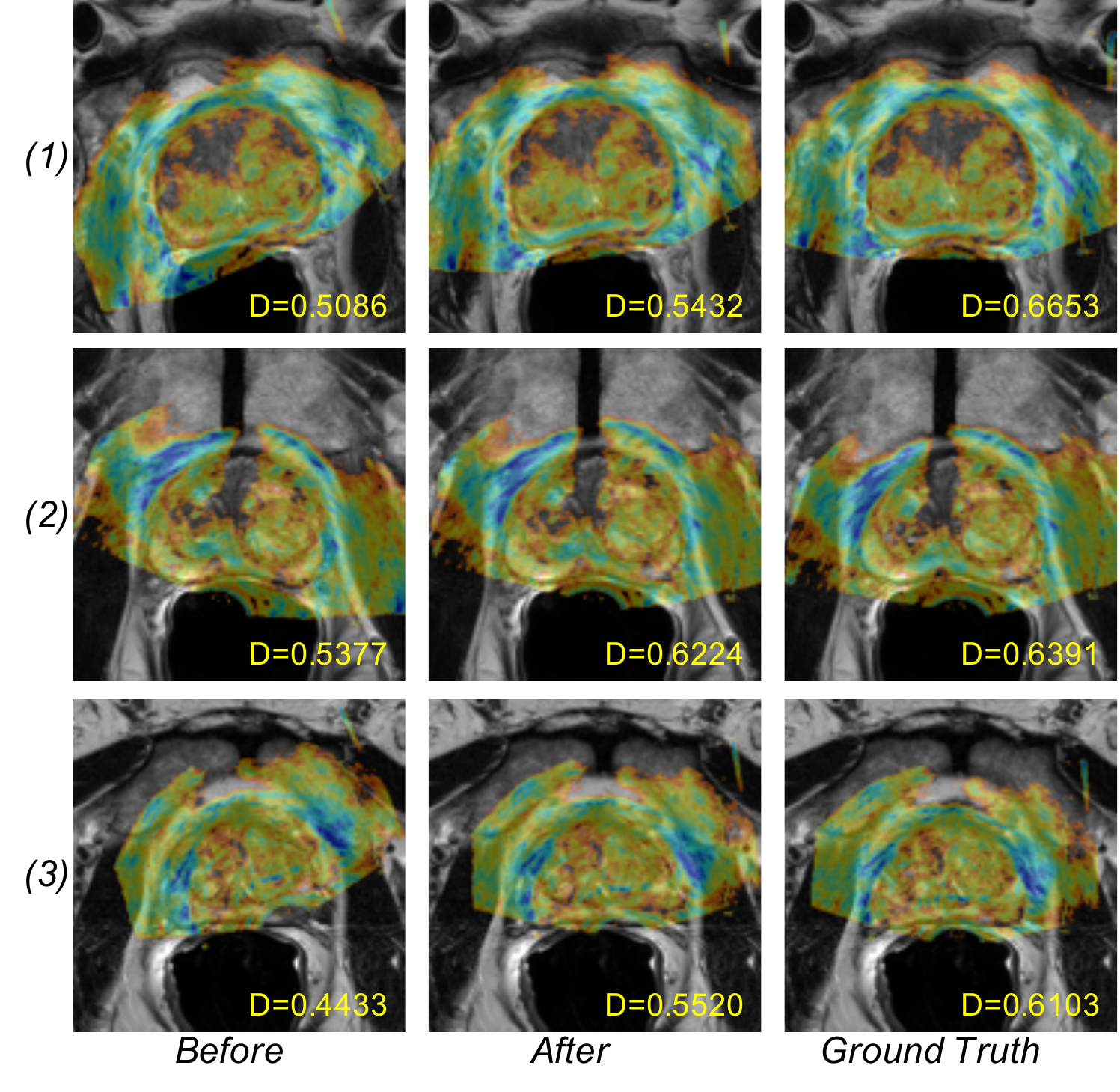}
\caption{\label{fig:example}Example registration results from 3 different cases. MR images are shown in gray level and corresponding TRUS images are superimposed in pseudo color. The columns from left to right are as follows. \textit{Left}: Images aligned under a randomly generated transform before registration; \textit{Middle}: Images aligned  using the generated transformation after registration; \textit{Right}: Images aligned using the manually performed registration by experts, which is considered as ground truth. The discriminator score for each pair of aligned images is shown in yellow at the lower right corner of the image.}
\end{figure}

\subsection{Experimental Results}

%\begin{table}[tb]
%\begin{center}
%	\caption{\label{tab:results}Experimental results.}
%	\begin{tabular}{l|c}
%		\hline \hline
%		Registration method & Registration accuracy (TRE) \\
%		\hline
%AIR-net WGAN & xx \\
%AIR-net WGAN + image difference & xx \\
%AIR-net WGAN + supervised & xx \\
%AIR-net WGAN + image difference + supervised & xx \\
%End-to-End unsupervised & ?? \\
%End-to-End supervised & ?? \\
%		\hline\hline
%	\end{tabular}
%\end{center}
%\end{table}
%
%The experimental results are given in Table~\ref{tab:results}.

With the trained networks, performance evaluation was then carried out. For each evaluation case, an initial transformation was randomly created in the same way as the training data by perturbing the ground truth transformation. The target registration error (TRE) and the discriminator scores (D-Scores) are then computed on the initial registration. The initial poorly aligned image pairs are input into the G-network for registration and a new set of transformation parameters are generated. The TRUS volume is then resampled by using the new registration and put together with the MR volume to form a new pair. TRE of the new registration will be computed and the new pair will also be fed into the D-network for scoring.

In our current experiment, we limit the randomly generated transformation to be in 2D, i.e. only rotation and translations in the axial view with 3 degrees of freedom. We are extending the method to more general scenarios.
Fig.~\ref{fig:example} first shows some example registration results. It can be seen that starting from some randomly perturbed registrations, the developed method was able to put the images back into alignment and get very close to the ground truth registration. The improved image alignment is also reflected by the D-Scores. As the registration quality improves, the D-scores also increase. This suggests that both the generator and discriminator networks are working effectively.

\begin{figure}[tbh!]
	\centering
	\subfloat[Target Registration Errors]{\includegraphics[width=.49\textwidth]{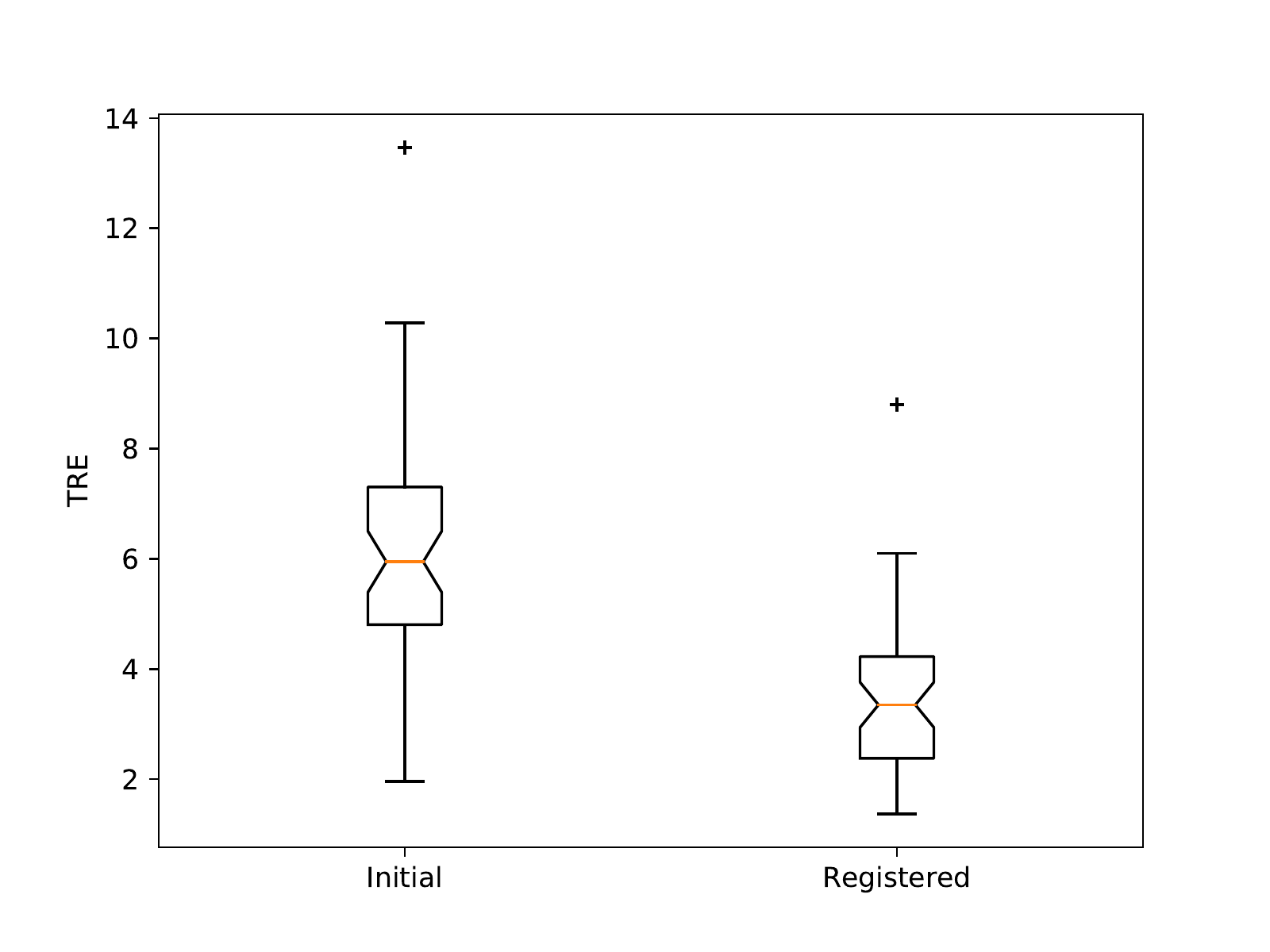}}
	\hfill
	\subfloat[D-Scores]{\includegraphics[width=.49\textwidth]{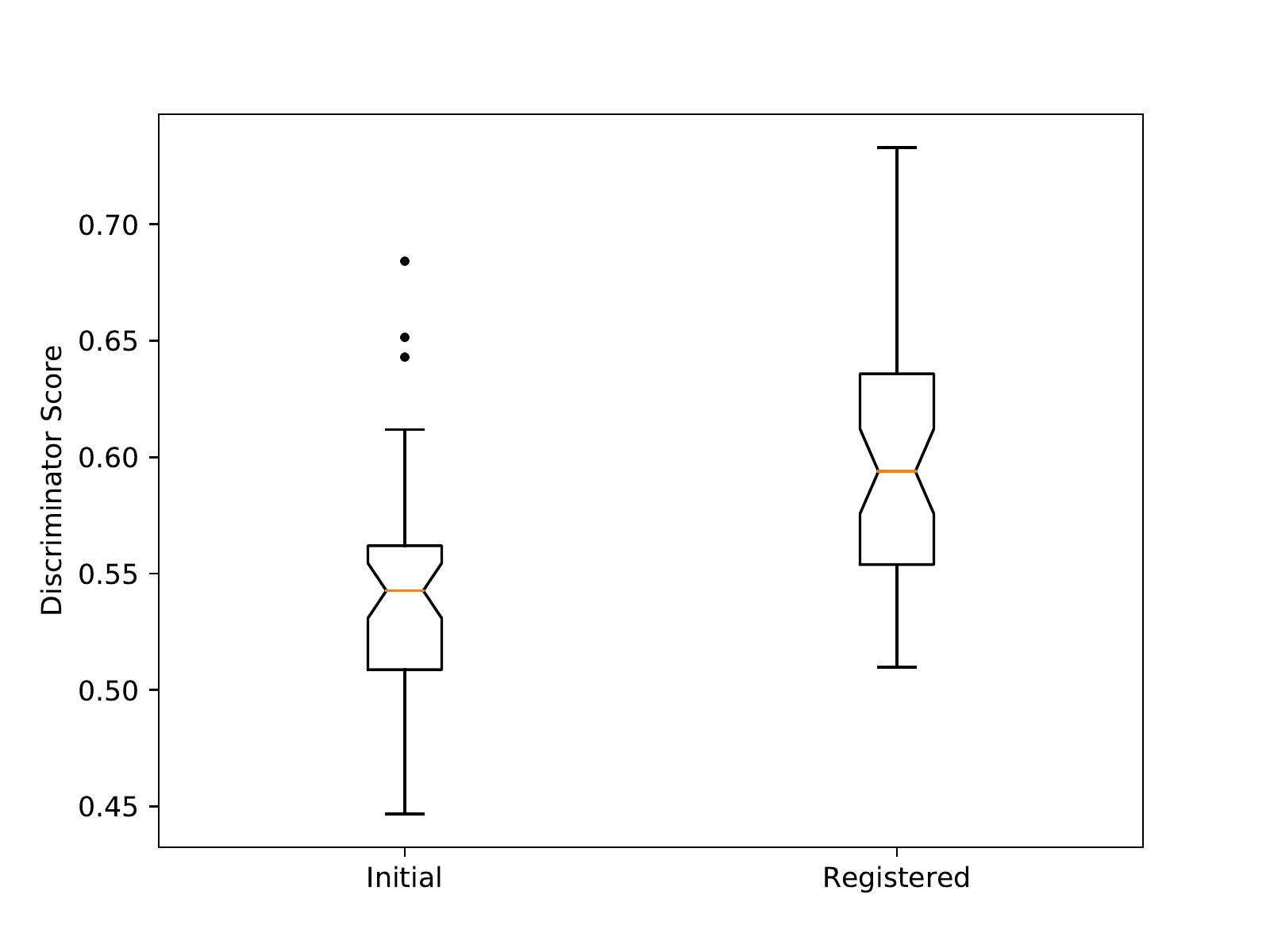}}
\caption{\label{fig:evaluation}Evaluation of the AIR-net based image registration performance  measured by TRE and D-Scores.}
\end{figure}

The registration performance of the developed AIR-net was then quantitatively evaluated and the results are given in Fig.~\ref{fig:evaluation}. The evaluation was performed using both TRE and D-Scores given by the D-network, respectively.  It can be seen from Fig.~\ref{fig:evaluation}(a) that the TRE dropped significantly ($p<$0.01) after registration, with mean TRE being decreased to 3.48mm from 6.11mm. in the same time, the D-scores are significantly ($p<$0.01) improved after image registration, which shows very good correlation with TRE. Therefore, the results demonstrate that the G-network is able to generate improved registration with significantly smaller registration error and the D-network is able to tell good registration from poor registration.

% =================================== %
\section{Conclusions}
\label{sec:conclusions}

In this paper, a new multi-modality image registration method of AIR-net based on the GAN framework is presented. To the best of our knowledge, this is the first work using GAN for multi-modality medical image registration. The proposed method provides not only a registration estimator, but also a quality evaluator in the same time, which can be used for quality check to detect potential registration failure. Being a major contribution of this work, it can be very useful in clinical practice to warn physicians about potential problems in image-fusion guided procedures. More evaluation will be performed in our future work against other state-of-the-art methods on registration performance.

% ----------------
\section*{Acknowledgment}

The authors would like to thank NVIDIA Corporation for the donation of the Titan Xp GPU used for this research.

% ======= references ======== %

%\bibliographystyle{splncs}
%\bibliography{registration_refs}

\end{document}